\documentclass{article}

\usepackage[utf8]{inputenc} 
\usepackage[T1]{fontenc}    
\usepackage{hyperref}       
\usepackage{url}            
\usepackage{booktabs}       
\usepackage{amsfonts}       
\usepackage{nicefrac}       
\usepackage{microtype, float, wrapfig}      
\usepackage{xcolor} 
\usepackage{bbm}

\usepackage[T1]{fontenc}
\usepackage{lmodern}

\usepackage{amsbsy,amssymb,amsmath,amsthm,latexsym,multicol,multirow,epsfig,color,subfigure,setspace,array,mathrsfs,bm,algpseudocode,algorithm}

\usepackage[stable]{footmisc}
\usepackage{graphicx}
\usepackage{enumitem}
\usepackage{authblk}

\usepackage[square,sort,comma,numbers]{natbib}

\newtheorem{theorem}{Theorem}[section]
\newtheorem{lemma}[theorem]{Lemma}

\newtheorem{corollary}[theorem]{Corollary}

\newcounter{assumption}[section]
\newenvironment{assumption}[1][]{\refstepcounter{assumption}\par\medskip
   \noindent \textbf{Assumption~\theassumption. #1} \rmfamily}{\medskip}

\newcounter{remark}[section]

\newcounter{example}[section]

\usepackage[T1]{fontenc}
\usepackage[utf8]{inputenc}
\usepackage{authblk}

\def\H{{\mathcal{H}}}

\def\SS{{\mathcal{S}}}
\def\YY{{\mathcal{Y}}}
\def\AA{{\mathcal{A}}}

\def\etahat{\widehat{\eta}}

\DeclareMathOperator{\Tr}{Tr}

\def\X{{\mathbf{X}}}
\def\K{{\mathbf{K}}}

\newcommand{\RR}{\mathbb R}

\def\T{\intercal}

\newcommand{\y}{\bm{y}}
\newcommand{\x}{\bm{x}}
\newcommand{\z}{\bm{z}}

\newcommand{\norm}[1]{\left\lVert#1\right\rVert}

\newcommand{\la}{{\langle}}
\newcommand{\ra}{{\rangle}}

\newcommand{\ex}{\mathbb{E}}

\newcommand{\bea}{\begin{eqnarray}}
\newcommand{\eea}{\end{eqnarray}}
\newcommand{\beaa}{\begin{eqnarray*}}
\newcommand{\eeaa}{\end{eqnarray*}}

\def\spacingset#1{\renewcommand{\baselinestretch}
  {#1}\small\normalsize} \spacingset{1} \allowdisplaybreaks

\begin{document}

\title{\bf Oversampling Divide-and-conquer for Response-skewed Kernel Ridge Regression}
  \author{Jingyi Zhang  \\
  	Tsinghua University, China \\
  	Xiaoxiao Sun \\ 
  	The University of Arizona}
\date{}
\maketitle

\begin{abstract}
  The divide-and-conquer method has been widely used for estimating large-scale kernel ridge regression estimates. Unfortunately, when the response variable is highly skewed, the divide-and-conquer kernel ridge regression (dacKRR) may overlook the underrepresented region and result in unacceptable results. We combine a novel response-adaptive partition strategy with the oversampling technique synergistically to overcome the limitation. Through the proposed novel algorithm, we allocate some carefully identified informative observations to multiple nodes (local processors). Although the oversampling technique has been widely used for addressing discrete label skewness, extending it to the dacKRR setting is nontrivial. We provide both theoretical and practical guidance on how to effectively over-sample the observations under the dacKRR setting. Furthermore, we show the proposed estimate has a smaller risk than that of the classical dacKRR estimate under mild conditions. Our theoretical findings are supported by both simulated and real-data analyses.
\end{abstract}

\section{INTRODUCTION}
We consider the problem of calculating large-scale kernel ridge regression (KRR) estimates in a nonparametric regression model. Although the theoretical properties of the KRR estimator are well-understood \citep{geer2000empirical,zhang2005learning,steinwart2009optimal}, in practice, the computation of KRR estimates may suffer from a large computational burden. In particular, for a sample of size $N$, it requires $O(N^3)$ computational time to calculate a KRR estimate using the standard approach, as will be detailed in Section~2. Such a computational cost is prohibitive when the sample size $N$ is considerable.
The divide-and-conquer approach has been implemented pervasively to alleviate such computational burden \citep{zhang2013divide,zhang2015divide,xu2016feasibility,xu2018divide,xu2019distributed}.
Such an approach randomly partitions the full sample into $k$ subsamples of equal sizes, then calculates a local estimate on an independent local processor (also called a local node) for each subsample. The local estimates are then averaged to obtain the global estimate.
The divide-and-conquer approach reduces the computational cost of calculating KRR estimates from $O(N^3)$ to $O(N^3/k^2)$.
Such savings may be substantial as $k$ grows. 

Despite algorithmic benefits, the success of the divide-and-conquer approach highly depends on the assumption that the subsamples can well represent the observed full sample. Nevertheless, this assumption cannot be guaranteed in many real-world applications, where the response variable $Y$ may have a highly skewed distribution.
Specifically, the random variable $Y$ has a highly skewed distribution $f_Y$ if $f_Y$ is nearly zero inside a large region $A_Y\subset\RR$. 
Problems of this type arise in high energy physics, Bayesian inference, financial research, biomedical research, environmental data, among others \citep{mcguinness1997statistical,afifi2007methods,haixiang2017learning}. 
In these applications, $Y \in A_Y$ are the responses that occur with reduced frequency.
However, such responses are often of more interest as they tend to have a more widespread impact. 
For example, $Y\in A_Y$ may represent a rare signal for seismic activity or stock market fluctuations.
Overlooking such signals could resulting in a substantial negative impact on society either economically or in terms of human casualties.

\begin{figure}[!ht]
    \begin{center}
        \begin{tabular}{c}
            \includegraphics[scale = .45]{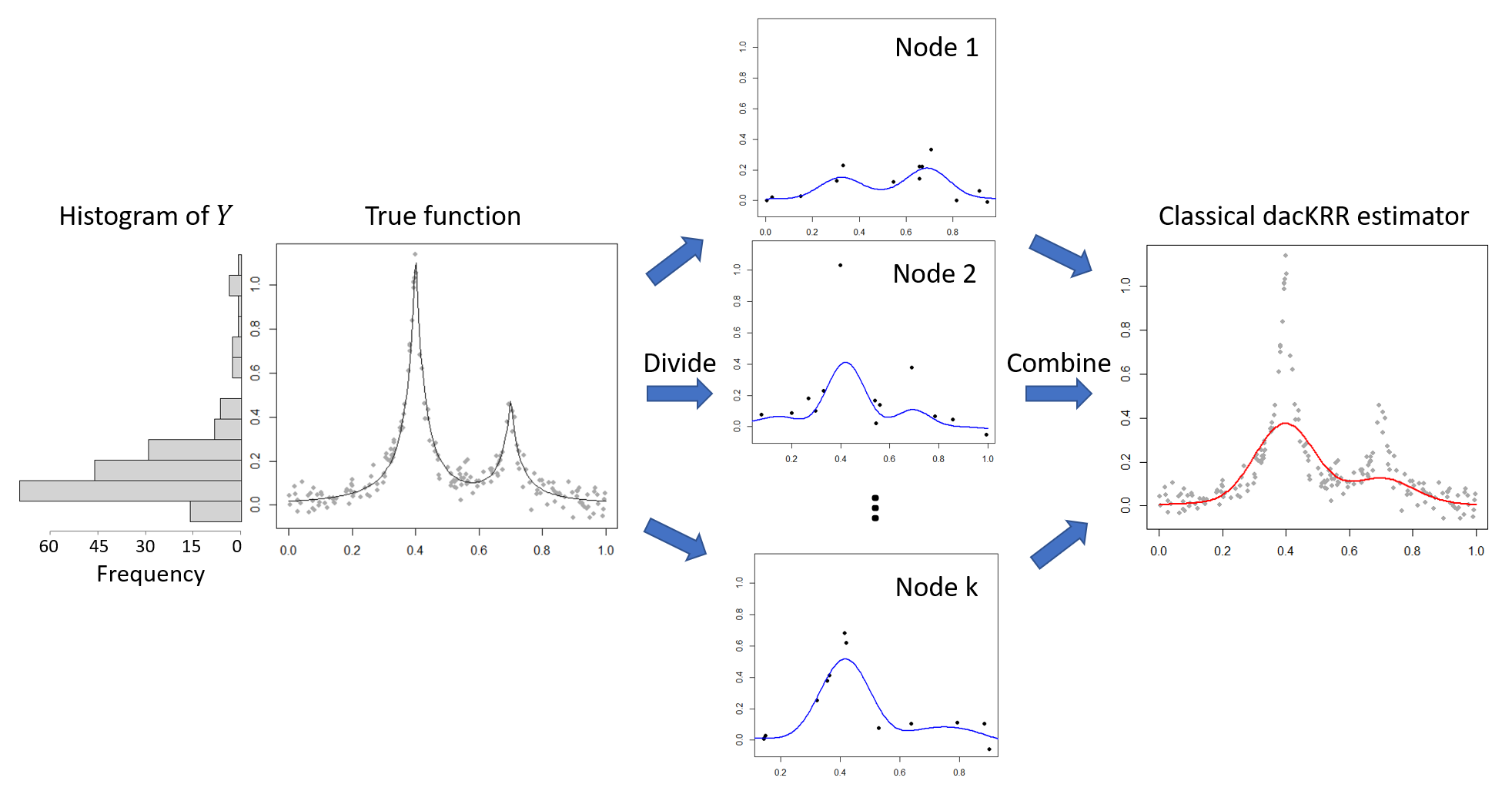}\\
        \end{tabular}
        \vspace{-5pt}
        \caption{An illustration of unacceptable results for the classical dacKRR estimate.}\label{fig1}
    \end{center}
     \vspace{-10pt}
\end{figure}

Recall that in the classical divide-and-conquer approach, a random subsample is processed on every local node.
Under the aforementioned rare-event scenarios, such a subsample could fail to have observations selected from the potentially informative region $A_Y$. 
The local estimate based on such a subsample thus is very likely to overlook the region $A_Y$.
Averaging these local estimators could lead to unreliable estimations and predictions over these informative regions.
A synthetic example in Fig.~\ref{fig1} illustrates the scenario that the response is highly skewed.
In this example, the one-dimensional sample (gray points) is uniformly generated from $[0,1]$.
The response variable $Y$ has a heavy-tailed distribution, as illustrated by the histogram.
The classical dacKRR method is used to estimate the true function (gray curve).
We observe that almost all the local estimates (blue curves) 
Averaging these local estimates thus results in a global estimate (red curve) with unacceptable estimation performance.
Such an observation is due to the fact that the subsample in each node overlooks the informative region over the distribution of $Y$ (the peaks) and thus fails to provide a successful global estimate.

\textbf{Our contributions.}
To combat the obstacles, we develop a novel response-adaptive partition approach with oversampling to obtain large-scale kernel ridge regression estimates, especially when the response is highly skewed.
Although the oversampling technique is widely used for addressing discrete label imbalance, extending such a technique to the continuous dacKRR setting is nontrivial.
We bridge the gap by providing both theoretical and practical guidance on how to effectively over-sample the observations. 
Different from the classical divide-and-conquer approach, such that each observation is allocated to only one local node, we propose to allocate some informative data points to multiple nodes.
Theoretically, we show the proposed estimates have a smaller risk than the classical dacKRR estimates under mild conditions.
Furthermore, we show the number of strata  $l$ of the response vector and the number of nodes $k$ regulate the trade-off between the computational time and the risk of the proposed estimator.
In particular, when $k$ is well-selected, a larger $l$ is associated with a longer computational time as well as a smaller risk.
Such results are novel to the best of our knowledge.
Our theoretical findings are supported by both simulated and real-data analyses.
In addition, we show the proposed method is not specific to the dacKRR method and has the potential to improve the estimation of other nonparametric regression estimators under the divide-and-conquer setting.

\section{PRELIMINARIES}
\textbf{Model setup.}
Let $\H$ be a reproducing kernel Hilbert space (RKHS).
Such a RKHS is induced by the reproducing kernel $K(\cdot, \cdot):\RR^d \times \RR^d \rightarrow \RR$, which is a symmetric nonnegative definite function, and an inner product $\langle\cdot,\cdot\rangle_\H$ satisfying
$\langle g(\cdot), K(\x,\cdot)\rangle_\H = g(\x)$ for any $g\in\H$.
The RKHS $\H$ is equipped with the norm $||g||_\H = (\langle g(\cdot),g(\cdot)\rangle)^{1/2}$.
The well-known Mercer's theorem states that, under some regularity conditions, the kernel function $K(\cdot, \cdot)$ can be written as $K(\x,\z) = \sum_{j=1}^\infty \mu_j\phi_j(\x)\phi_j(\z)$, where $\{\mu_j\}_{j=1}^\infty \ge 0$ is a sequence of decreasing eigenvalues and $\{\phi_j(\cdot)\}_{j=1}^\infty$ is a family of orthonormal basis functions.
The smoothness of a function $g\in\H$ is characterized by the decaying rate of the eigenvalues $\{\mu_j\}_{j=1}^\infty$.
The major types of such decaying rates include the finite rank type, the exponentially decaying type, and the polynomially decaying type.
The representative kernel function of these three major types are the polynomial kernel $K(\x,\z) = (1+\x^\T\z)^r$ ($r$ is an integer), the Gaussian kernel $K(\x,\z) = \mbox{exp}(-||\x-\z||^2/\sigma^2)$ ($\sigma>0$ is the scale parameter and $||\cdot||$ is the Euclidean norm), and the kernel $K(x,z) = 1+\mbox{min}(x,z)$, respectively.
We refer to \cite{hastie2009elements,shawe2004kernel} for more details.


Consider the nonparametric regression model
\begin{equation}\label{model1}
y_i=\eta(\x_i)+\epsilon_i, \quad i=1,\ldots,N,
\end{equation}
where $y_i\in \mathbb{R}$ is the response, $\x_i\in\mathbb{R}^d$ is the predictors, $\eta$ is the unknown function to be estimated, and $\{\epsilon_i\}_{i=1}^N$ are i.i.d. normal random errors with zero mean and unknown variance $\sigma^2 < \infty$.
The kernel ridge regression estimator aims to find a projection of $\eta$ into the RKHS $\H$. Such an estimator $\etahat$ can be written as
\begin{eqnarray}\label{eqn1}
\etahat =\underset{\eta\in\H}{\mbox{argmin}}\left\{ \frac{1}{N}\sum_{i=1}^N\{y_i-\eta(\x_i)\}^2+\lambda ||\eta||_\H^2\right\}.
\end{eqnarray}
Here, the regularization parameter $\lambda$ controls the trade-off between the goodness of fit of $\etahat$ and the smoothness of it. 
A penalized least squares framework analogous to Equation~(\ref{eqn1}) has been extensively studied in the literature of regression splines and smoothing splines \citep{wahba1990spline,hastie1996pseudosplines,luo1997hybrid,he2001data,gu2002penalized,ruppert2002selecting,zhang2004variable,sklar2013nonparametric,yuan2013adaptive,ma2015efficient,zhang2018smoothing,meng2020more}.

The well-known representer theorem \cite{wahba1990spline} states that the minimizer of Equation~(\ref{eqn1}) in the RKHS $\H$ takes the form 
$\etahat(\bm{x})=\sum_{i=1}^N \beta_i K(\bm{x}_i,\bm{x}).$
Let $\y=(y_1,\ldots,y_N)^\T$ be the response vector, $\bm{\beta}=(\beta_1,\ldots,\beta_N)^\T$ be the coefficient vector and $\K$ be the kernel matrix such that the $(i,j)$-th element equals $K(\x_i,\x_j)$.
These coefficient vector $\bm{\beta}$ can be estimated through solving the minimization problem as follows,
\begin{equation}\label{eqn4}
\widehat{\bm{\beta}}=
\underset{\bm{\beta}\in\RR^N}{\mathrm{argmin}}\frac{1}{N}(\y-\K\bm{\beta})^\T(\y-\K\bm{\beta})+\lambda \bm{\beta}^\T \K\bm{\beta}.
\end{equation} 
It is known that the solution of such a minimization problem has a closed form
\begin{equation}\label{eqn5}
\widehat{\bm{\beta}} = 
(\K+N\lambda \mathbf{I}_N)^{-1}\y,
\end{equation} 
where the regularization parameter $\lambda$ can be selected based on the cross-validation technique, the Mallows's Cp method \citep{mallows2000some}, or the generalized cross-validation (GCV) criterion \citep{wahba1978smoothing}.

Although the solution of the minimization problem~(\ref{eqn4}) has a closed-form, the computational cost for calculating the solution using Equation~(\ref{eqn5}) is of the order $O(N^3)$, which is prohibitive when the sample size $N$ is considerable.
In this paper, we focus on the divide-and-conquer approach for alleviating such a computational burden.
In recent decades, there also exist a large number of studies that aim to develop low-rank matrix approximation methods to accelerate the calculation of kernel ridge regression estimates \citep{rahimi2008random,wang2015practical,rudi2015less,mahoney2016lecture,musco2017recursive}.
These approaches are beyond the scope of this paper.
In practice, one can combine the divide-and-conquer approach with the aforementioned methods to further accelerate the calculation.

\textbf{Background of the divide-and-conquer approach.}
The classical divide-and-conquer approach is easy to describe. 
Rather than solving the minimization problem~(\ref{eqn4}) using the full sample, the divide-and-conquer approach randomly partitions the full sample into $k$ subsamples of equal sizes. 
Each subsample is then allocated to an independent local node. 
Next, $k$ minimization problems are solved independently on each local node based on the corresponding subsamples.
The final estimate is simply the average of all the local estimates.
The divide-and-conquer approach has proven to be effective in linear models \citep{chen2014data,lu2016nonparametric}, partially linear models \citep{zhao2016partially}, nonparametric regression models \citep{zhang2013divide,zhang2015divide,lin2017distributed,shang2017computational,guo2017learning}, principal component analysis \citep{wu2018review,fan2019distributed}, matrix factorization \citep{mackey2011divide},  among others.

\begin{algorithm}
\caption{Classical divide-and-conquer kernel ridge regression}
\label{algo1}
\begin{algorithmic}
\State {\bfseries Input:} The training set $\{(\x_i, y_i)\}_{i=1}^N$; the number of nodes $k$
   \State \textbf{Step 1:} Randomly and evenly partition the sample $\{(\x_1, y_1), \ldots,(\x_N,y_N)\}$ into $k$ disjoint subsamples, denoted by $\SS_1,\ldots,\SS_k$. Let $|\SS_i|$ be the number of observations in $\SS_i$.
   \State \textbf{Step 2: } 
   \For{$j$ in $1,\ldots,k$}
   \State calculate the local kernel ridge regression estimate on the $j$-th local node
   $$\hat{\eta}_j = \underset{\eta\in\H}{\mathrm{argmin}}\{ \frac{1}{|\SS_j|}\sum_{(\x,y)\in\SS_j}(y-\eta(\x))^2+\lambda_j ||\eta||_\H^2\}.$$
   \EndFor
   \State \textbf{Output: } Combine each local estimates to obtain the final estimate
   $\widetilde{\eta} = \frac{1}{k} \sum_{j=1}^k \hat{\eta}_j.$
\end{algorithmic}
\end{algorithm}

Algorithm~\ref{algo1} summarizes the classical divide-and-conquer method under the kernel ridge regression setting.
Such an algorithm reduces the computational cost for the estimation from $O(N^3)$ to $O(N^3/k^2)$. 
The savings may be substantial as $k$ grows.

One difficulty in Algorithm~1 is how to choose the
regularization parameter $\lambda_j$s. 
A natural way to determine the size of these regularization parameters is to utilize the standard approaches, e.g., the Mallows's Cp criterion or the (GCV) criterion, based only on the local subsample $\SS_j$. It is known that such a simple strategy may lead to a global estimate that suffers from suboptimal performance \citep{zhang2015divide,xu2018divide}.
To overcome the challenges, there has been a large number of studies dedicated to developing more effective methods of selecting the local regularization parameters in the recent decade.
For example, \cite{zhang2015divide} proposed to select the regularization parameter $\lambda_j$ according to the order of the entire sample size $N$ instead of the subsample size $N/k$. 
Recently, \cite{xu2018divide} proposed the distributed GCV method to select a global optimal regularization parameter for nonparametric regression under the classical divide-and-conquer setting.

\section{MOTIVATION AND METHODOLOGY}

\textbf{Motivation.}
To motivate the development of the proposed method, we first re-examine the oversampling strategy. 
Such a strategy is well-known in imbalanced data analysis, where the labels can be viewed as skewed discrete response variables. \citep{japkowicz2002class,he2009learning,krawczyk2016learning}.
Classical statistical and machine learning algorithms assume that the number of observations in each class is roughly at the same scale. 
In many real-world cases, however, such an assumption may not hold, resulting in imbalanced data.
Imbalanced data pose a difficulty for classical learning algorithms, as they will be biassed towards the majority group. 
Despite the rareness, the minority class is usually more important from the data mining perspective, as it may carry useful and important information. 
An effective classification algorithm thus should take such unbalances into account.
To tackle the imbalanced data, the oversampling strategy supplements the training set with multiple copies of the minority classes and keeps all the observations in the majority classes to make the whole dataset suitable for a standard learning algorithm.
Such strategy has proven to be effective for achieving more robust results \citep{japkowicz2002class,he2009learning,krawczyk2016learning}.

Intuitively, the data with a continuous skewed response can be considered as imbalanced data.
In particular, if we divide the range of the skewed response $\{y_i\}_{i=1}^N$ into $l$ equally-spaced disjoint intervals, denoted by $\YY_1,\ldots,\YY_l$, then there will be a disproportionate ratio of observations in each interval.
The intervals that contain more observations can be considered as the majority classes, and the ones that contain fewer observations can be considered as the minority classes.
Recall that the classical divide-and-conquer approach utilizes a simple random subsample from the full sample to calculate the local estimate. Therefore, when the response variable is highly skewed, such a subsample could easily overlook the observations respecting the minority classes, resulting in unpleasant estimates.

\textbf{Main algorithm.}
Inspired by the oversampling strategy, we propose to supplement the training data with multiple copies of the minority classes before applying the divide-and-conquer approach.
In addition, analogous to the downsampling strategy, the subsample in each node should keep most of the observations from the minority classes.
Let $|\SS|$ be the number of observations in the set $\SS$.
Let $[\cdot]$ be the rounding function that rounds down the number to its nearest integer.
The proposed method, called oversampling divide-and-conquer kernel ridge regression, is summarized in Algorithm~\ref{algo2}.

\begin{figure}[!ht]
    \begin{center}
        \begin{tabular}{c}
            \includegraphics[scale = .55]{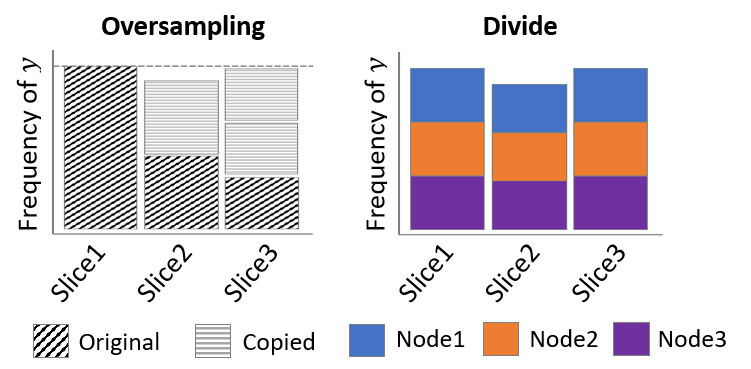}\\
        \end{tabular}
        \vspace{-5pt}
        \caption{Illustration of Algorithm~\ref{algo2}.}\label{fig2}
    \end{center}
     \vspace{-10pt}
\end{figure}


We use a toy example to illustrate Algorithm~\ref{algo2} in Fig.~\ref{fig2}.
Suppose there are three nodes ($k=3$), and the histogram of the response is divided into three slices ($l=3$).
The left panel of Fig.~\ref{fig2} illustrates the oversampling process.
In particular, the observations are duplicated by certain times, such that the total number of observations respecting each slice is roughly equal to each other.
The original observations and the duplicated observations are marked by tilted lines and horizontal lines, respectively.
The right panel of Fig.~\ref{fig2} shows the next step.
For each slice, the observations within such a slice are then randomly and evenly allocated to each of the nodes.
The observations allocated to different nodes are marked by blue, red and purple, respectively.
Finally, similar to the classical dacKRR approach, the local estimates are calculated and then are averaged to obtain the global estimate.


\begin{algorithm}
\caption{Oversampling divide-and-conquer kernel ridge regression}
\label{algo2}
\begin{algorithmic}
\State {\bfseries Input:} The training set $\{(\x_i, y_i)\}_{i=1}^N$; the number of nodes $k$, the number of slices $l$
   \State \textbf{Step 1:} Divide the range of the responses $\{y_i\}_{i=1}^N$ into $l$ disjoint equally-spaced intervals, denoted by $\YY_1,\ldots,\YY_l$. Without lose of generality, we assume $|\YY_1|\geq\cdots\geq|\YY_l|$.
   \State \textbf{Step 2: }
   \For{$j$ in $1,\ldots,l$}
   \State Let $\YY_j^*$ be the set that contains $[|\YY_1|/|\YY_j|]$ copies of all the elements in $\YY_j$. Randomly and evenly partition the set $\{(\x, y)|y\in\YY_j^*\}$ into $k$ disjoint subsamples, denoted by $\AA_{1,j},\ldots,\AA_{k,j}$.  
   \EndFor
   \State \textbf{Step 3: } 
   \For{$i$ in $1,\ldots,k$}
   \State The subsample respecting the $i$-th local node is $\SS_i = \mathcal{D}\{\AA_{i,1}\cup\cdots\cup\AA_{i,l}\}$, where $\mathcal{D}\{\cdot\}$ denotes de-duplicating. Calculate the local estimate on the $i$-th local node 
   $$\hat{\eta}_i = \underset{\eta\in\H}{\mathrm{argmin}}\{ \frac{1}{|\SS_i|}\sum_{(\x,y)\in\SS_i}(y-\eta(\x))^2+\lambda_i ||\eta||_\H^2\}.$$
   \EndFor
   \State \textbf{Output:} Combine each local estimates to obtain the final estimate
   $\bar{\eta} = \frac{1}{k} \sum_{i=1}^k \hat{\eta}_i.$
\end{algorithmic}
\end{algorithm}

\textbf{Implementation details and the computational cost.}
Analogous to Algorithm~\ref{algo1}, a difficulty in Algorithm~\ref{algo2} is how to choose the regularization parameters $\lambda_i$'s.
Throughout this paper, we opt to use the same approach to select $\lambda_i$'s as the one proposed in \cite{zhang2015divide}.
The number of slices $l$ can be set as a constant or be determined by Scott's rule; see \cite{scott2015multivariate} for more details.
Our empirical studies indicate that the performance of the proposed algorithm is robust to a wide range of choices of $l$. In addition, to avoid the data dependency within a node raised by oversampling, we conduct a ``de-duplicating'' process before calculating the local estimates.

Consider the total number of observations in the training set after the oversampling and de-duplicating processes.
Let $\tilde{N}$ denote such a number.
When the response variable has a roughly uniform distribution, few observations would be copied, and thus one has $\tilde{N}=O(N)$. In such a case, the computational cost of Algorithm~\ref{algo2} is at the same order as that of the classical divide-and-conquer method.
In the most extreme case that the most majority class $\YY_1$ contains almost all the observations, it can be shown that $\tilde{N}\le lN$.
Therefore, the computational cost of Algorithm~\ref{algo2} is at the order of $O(l^3N^3/k^2)$.
Consequently, when $l$ is a constant, again, the computational cost of Algorithm~\ref{algo2} is at the same order of the classical divide-and-conquer method.
Otherwise, when $l$ and $k$ go to infinity, for example, when $l$ is determined by Scott's rule \citep{scott2015multivariate} ( $l=O(N^{1/3})$), and $k = O(l)$, the computational cost of Algorithm~\ref{algo2} becomes $O(N^4/k^2)$.
However, such a higher computational cost is associated with theoretical benefits, as will be detailed in the next section.

\section{THEORETICAL RESULTS}
\textbf{Main theorem.}
In this section, we obtain the upper bounds on the risk of the proposed estimator $\bar{\eta}$ and establish the main result in Theorem \ref{thm:bound}. 
We show our bounds contain a smaller asymptotic estimation variance term, compared to the bounds of the classical dacKRR estimator \citep{zhang2015divide,liu2018many}. The following assumptions are required to bound the terms in Theorem \ref{thm:bound}. Assumption~\ref{as:1} guarantees the spectral decomposition of the kernel via Mercer's theorem.
Assumption \ref{as:2} is a regularity assumption on $\eta_0$. These two assumptions are fairly standard. Assumption \ref{as:3} requires that the basis functions are uniformly bounded. 
\cite{zhang2015divide} showed that one could easily verify this assumption for a given kernel $K$. For instance, this assumption holds for many classical kernel functions, e.g., the Gaussian kernel.

\begin{assumption}
The reproducing kernel $K$ is symmetric, positive definite, and square integrable. 
\label{as:1}
\vspace{-10pt}
\end{assumption}
\begin{assumption}
The underlying function $\eta_0 \in \mathcal{H}$. 
\label{as:2}
\vspace{-10pt}
\end{assumption}
\begin{assumption}
There exists some $\omega^2 > 0$ such that 
$
    \sup_{\x \in \mathcal{X}} \sum_{j=1}^{\infty} u_j \phi_j(\x)^2 \leq \omega^2 < \infty
$.
\label{as:3}
\end{assumption}

Let $\{\tilde{\X}_i, \tilde{\bm{y}}_i\}$ be the subset with size of $\tilde{n}$ allocated to the $i$th node after de-duplicating, $i=1,\ldots, k$. We assume they are drawn i.i.d from an unknown probability measure $\mathcal{P}$. Let $\mathcal{P}_{X}$ denote marginal distribution of $\tilde{\X}_i$. 
For brevity, we further assume the sample size for each node is the same. 
The bounds on the risk of the proposed estimator $\bar{\eta}$ are given in Theorem~\ref{thm:bound}, followed by the main corollary.  

\begin{theorem}
Let $d_{\lambda}=\sum_{j\ge 1} (1+\lambda/\mu_j)^{-1}$ be the the effective dimensionality of the kernel $K$ \citep{zhang2005learning}. Under Assumption \ref{as:1}-\ref{as:3}, the risk of the proposed estimator $\bar{\eta}$ is bounded by 
\begin{equation*}
  \begin{split}
    \ex \{\norm{\bar{\eta}-\eta_0}^2_{\mathcal{P}_X}\}  &\leq 8\lambda \norm{\eta_0}^2_{\mathcal{H}} + \frac{4d_{\lambda}\sigma^2}{\tilde{n}k}+ 4d_{\lambda}\\
    &\times(\mu_1^2 \norm{\eta_0}^2_{\mathcal{H}}+\frac{\omega^2\sigma^2}{\lambda \tilde{n}k})\exp\bigg(-\frac{3\lambda \tilde{n}}{28\omega^2}\bigg).
   \end{split}
\end{equation*}
\label{thm:bound}
\end{theorem}

\begin{corollary}
Assume the first two terms of the bounds in Theorem \ref{thm:bound} are dominant, we have
\begin{equation}\label{eqn11}
    \ex \{\norm{\bar{\eta}-\eta_0}^2_{\mathcal{P}_X}\} = O(1)\bigg\{\underbrace{\lambda \norm{\eta_0}^2_{\mathcal{H}}}_{\text{Squared bias}} + \underbrace{\frac{\sigma^2d_{\lambda}}{\tilde{N}}}_{\text{Variance}}\bigg\}.
\end{equation}
\label{coro}
\end{corollary}
\vspace{-10pt}
With the properly chosen regularization parameter, most of the commonly used kernels, e.g., the kernels with polynomial eigendecay, can satisfy this assumption for Corollary \ref{coro} \citep{gu2013smoothing}. More discussions can be found in the Supplementary Material.     
The squared bias term for the classical dacKRR estimator is at the order of $O(\lambda \norm{\eta_0}^2_{\mathcal{H}})$, which is the same as the one in Equation~(\ref{eqn11}) \citep{zhang2015divide}.
However, different from the variance term in Equation~(\ref{eqn11}), which equals $O(\sigma^2d_\lambda/\tilde{N})$, the variance term for the classical dacKRR estimator is at the order of $O(\sigma^2 d_{\lambda}/N)$. Our rate depends on the eigenvalues $\{\mu_j^2\}$ instead of the trace of kernel derived in \cite{zhang2015divide}. This rate is tighter than that in \cite{BAUER200752} and is considered as the minimax rate.

\textbf{Practical guidance on the selection of the parameter $l$ and $k$.}
We now compare the risk of the classical dacKRR estimator with the proposed one under three different scenarios. Furthermore, we provide some practical guidance on the selection of the number of slides $l$ and the number of nodes $k$. We denote the total sample size after oversampling as $\tilde{N}^*$.

\begin{figure}[!ht]
    \begin{center}
        \begin{tabular}{c}
            \includegraphics[scale = .6]{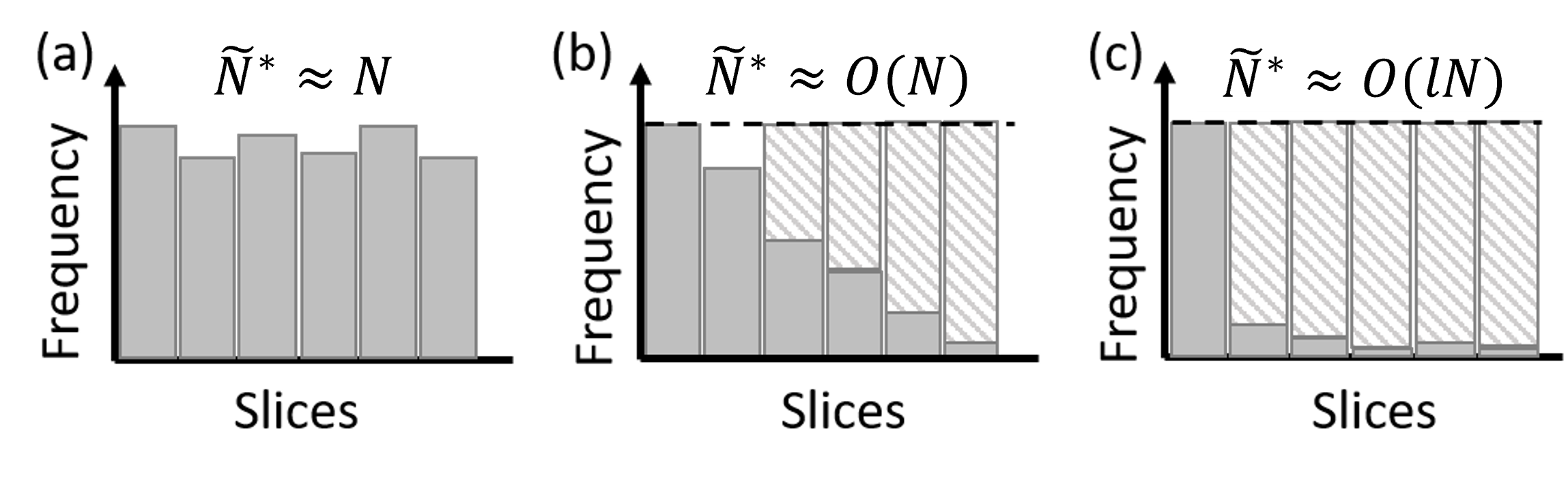}\\
        \end{tabular}
         \vspace{-10pt}
        \caption{Three different scenarios for the ``skewness''. The solid bars are original data and the lined ones are over-sampled data. }\label{skewness}
    \end{center}
     \vspace{-10pt}
\end{figure}

First, consider the scenario that $N\approx \tilde{N}^*$ as shown in Fig.~\ref{skewness}(a). In this case, oversampling is unnecessary, thus the proposed estimate $\bar{\eta}$ has the same risk as the classical dacKRR estimator.

Second, consider the scenario that the response variable has a slightly skewed distribution, i.e., one has $\tilde{N}^* = O(N)$ as shown in Fig.~\ref{skewness}(b). Under this scenario, the total sample size after de-duplicating is also in the same order of the original sample size, i.e. $\tilde{N} = O(N)$. 
Corollary~\ref{coro} indicates the proposed estimate $\bar{\eta}$ has the same risk as the classical dacKRR estimator.

Third, consider the scenario that the response variable has a highly skewed distribution. Under this scenario, the sample size in the ``majorest'' slide is in the same order of $N$, and the sample size in the ``minorest'' slide, denoted as $N_l$, is in the order of $o(N)$. In these cases, one has $\tilde{N}^* = O(lN)$ 
We then discuss under two cases, (1) when $N_l = o(N/l)$, especially, when $l$ is a constant; and (2) when $N_l = O(N/l)$. In case (1), the sample size $N$ after de-duplicating is in the same order of $N$. The risk of $\bar{\eta}$ then has the same order as the classical dacKRR estimator. In case (2), notice that $N_l = o(N)$, this indicates that $l$ should go to infinity as $N\to\infty$. In this scenario, we consider $k = O(N/N_l)$ i.e., $k = O(l)$, then we have $\tilde{N} = O(\tilde{N}^*)$. Thus equation~(\ref{eqn11}) becomes
\begin{equation}\label{eqn12}
    \ex \{\norm{\bar{\eta}-\eta_0}^2_{\mathcal{P}_X}\} = O(1)\bigg\{\lambda \norm{\eta_0}^2_{\mathcal{H}} + \frac{\sigma^2d_{\lambda}}{lN}\bigg\}.
\end{equation}
For example, when $l$ is determined by Scott's rule, one has $l=O(N^{1/3})$, and we choose $k=O(l)$.
Equation~(\ref{eqn12}) indicates the proposed estimator has a smaller risk than the classical dacKRR estimator. 
Corollary~\ref{coro} indicates that the parameters $l$ and $k$ regulate the trade-off between the computational time and the risk.
Specifically, when $k$ is well-selected, a larger $l$ is associated with a longer computational time as well as a smaller risk.

\section{SIMULATION RESULTS}
To show the effectiveness of the proposed estimator, we compared it with the classical dacKRR estimator in terms of the mean squared error (MSE).
We calculated the MSE for each of the estimators based on 100 replicates.
In particular, $\mbox{MSE}=\sum_{i=1}^{100}\|\hat{\eta}^{(i)} - \eta_0\|^2/100)$, where $\eta_0$ is the true function and $\hat{\eta}^{(i)}$ represents the estimator in the $i$th replication, respectively. 
Through all the experiments in this section, we set the number of nodes $k=100$. 
Gaussian kernel function was used for the kernel ridge regression. 
We followed the procedure in \cite{zhang2015divide} to select the bandwidth for the kernel and the regularization parameters.
For the proposed method, we divided the range of $\{y_i\}_{i=1}^N$ into $l$ slices according to Scott's rule \citep{scott2015multivariate}. 

\begin{figure}[!ht]
    \begin{center}
        \begin{tabular}{l}
            \includegraphics[scale = .45]{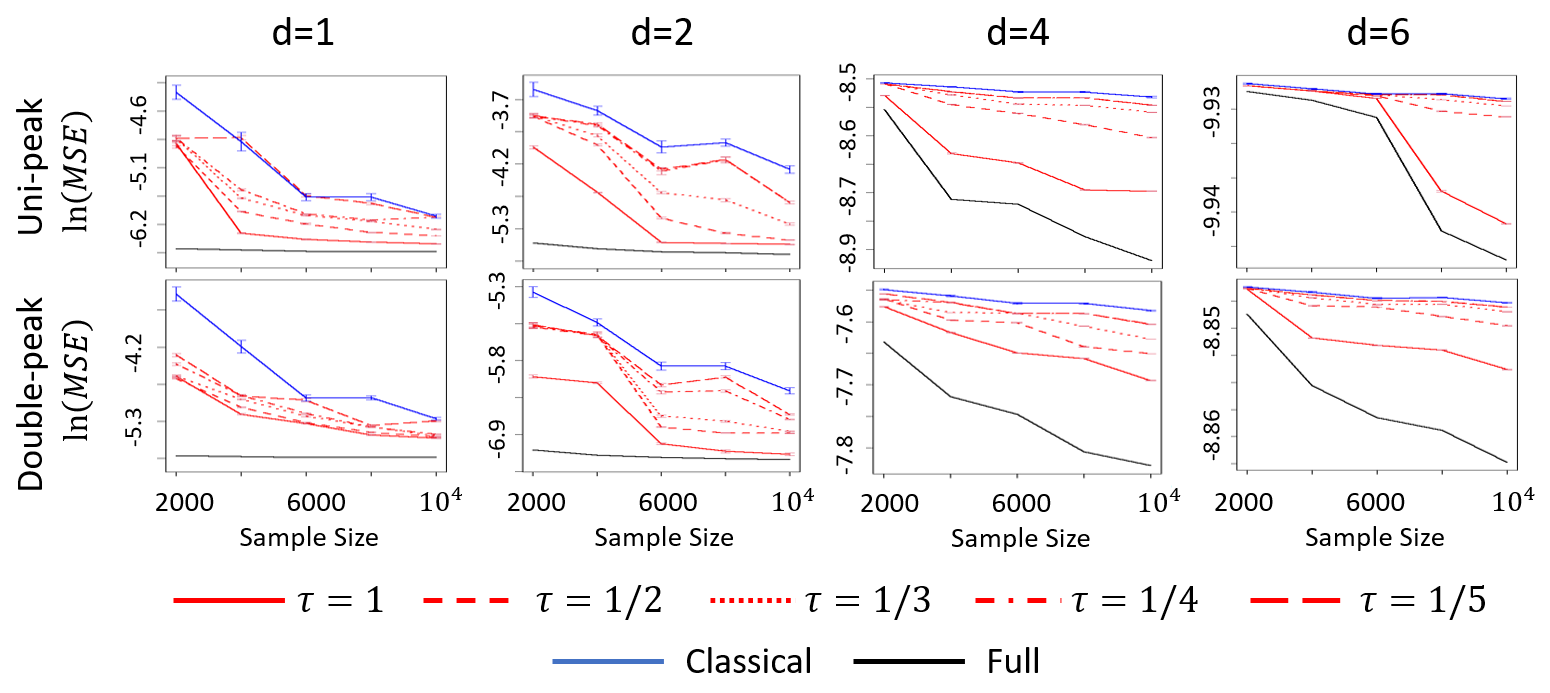}\\
        \end{tabular}
         \vspace{-5pt}
        \caption{Comparison of different estimators. Each row represents a different true function and each column represents a different $d$.}\label{simu_result}
    \end{center}
     \vspace{-10pt}
\end{figure}

\begin{figure}[!ht]
    \begin{center}
        \begin{tabular}{l}
            \includegraphics[scale = .4]{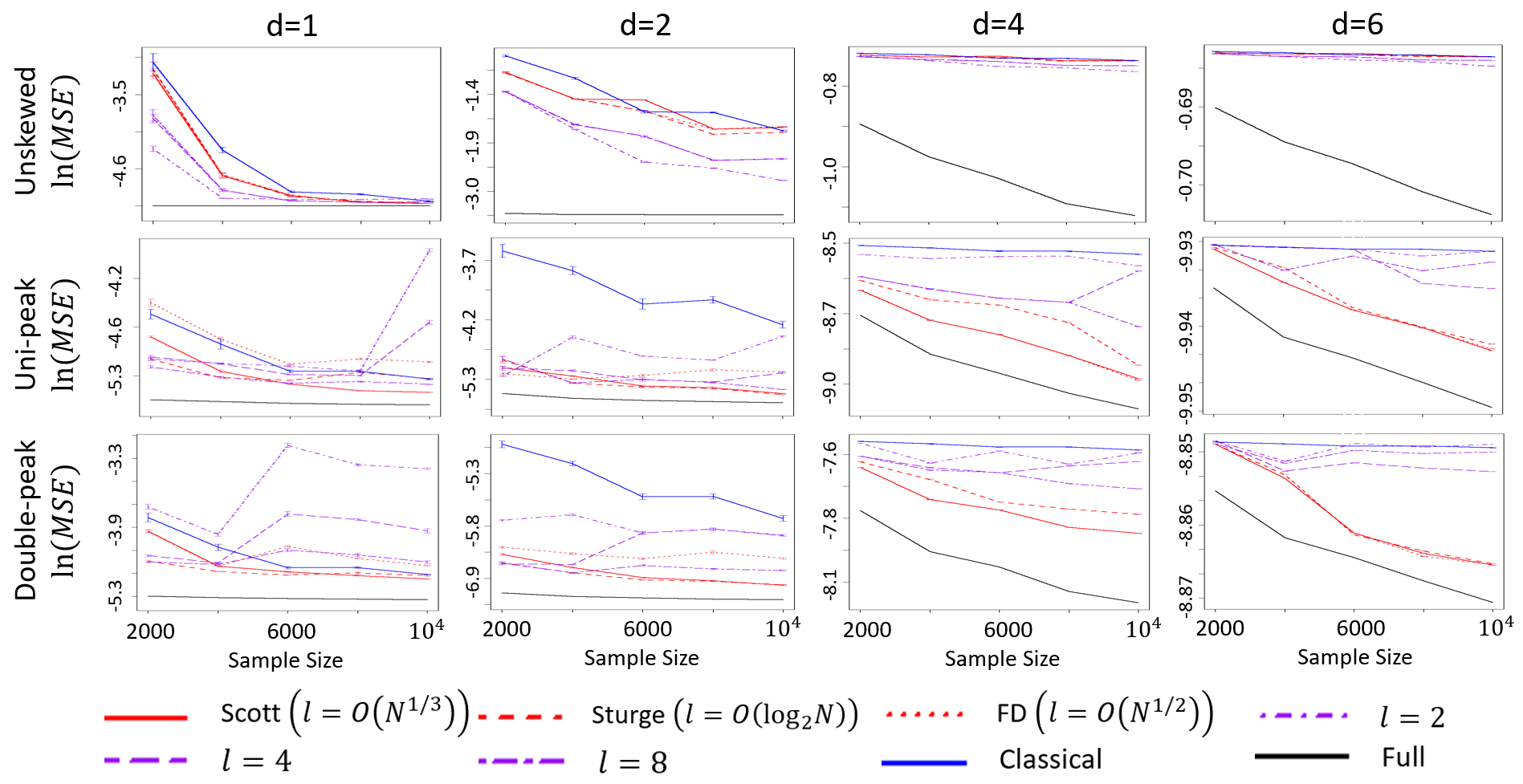}\\
        \end{tabular}
         \vspace{-5pt}
        \caption{Comparison for different choices of $l$'s.}\label{simu_breaks}
    \end{center}
     \vspace{-10pt}
\end{figure}

Recall that in Algorithm~2, each observation is copied for certain times, such that the total number of observations in each slice is roughly equal to each other.
A natural question is how does the number of such copies affects the performance of the proposed method.
In other words, if a fewer number of copies are supplemented to the training data, would the proposed method perform better or worse?
To answer this question, we induced a constant $\tau$ to control the oversampling size. 
Specifically, for the $j$th slice, $j=1,\ldots, l$, we duplicated the observations within such a slice for $([\tau|\YY_1|/|\YY_j|]-1)$ times, instead of $([|\YY_1|/|\YY_j|]-1)$ times in Algorithm~\ref{algo2}.
We set $\tau =\{1, 1/2, 1/3, 1/4, 1/5\}$.
This procedure is equivalent to Algorithm~\ref{algo2} when $\tau=1$.

Let $\bm{1}^d$ be the $d$-dimensional vector, such that all elements are equal to $1$.
We considered a function that has a highly skewed response, i.e.,
\begin{equation*}
    g(\bm{x}, \bm{c}) = \frac{0.1}{\|\bm{x}-\bm{c}\| +0.05}\sin\left(\frac{0.01\pi}{\|\bm{x}-\bm{c}\|+0.05}\right),
\end{equation*}
where $\bm{x}\in [0,1]^d$, and $\|\cdot\|$ represents the $\mathbb{L}_2$ norm.
We then simulated the data from Model~\eqref{model1} with $N=\{2,4,6,8,10\}\times10^3$, $d=1,2,4,6$, and two different regression function $\eta_0$'s,

{\textbf{Uni-peak}}: $\eta_0(\bm{x}) = g(\bm{x}, \bm{c})$, with $\bm{c} = 0.4\times\bm{1}^d$;

{\textbf{Double-peak}}: $\eta_0(\bm{x}) = g(\bm{x}, \bm{c}_1) + 0.4g(\bm{x}, \bm{c}_2)$, with $\bm{c}_1 = 0.4\times\bm{1}^d$ and $\bm{c}_2 = 0.7\times\bm{1}^d$.

Figure~\ref{simu_result} shows the MSE versus different sample size $N$ under various settings.
Each column represents a different $d$, and each row represents a different $\eta_0$.
The classical dacKRR estimator and the full sample KRR estimator are labeled as blue lines and black lines, respectively.
The proposed estimators are labeled as red lines, and the proposed method with $\tau=1$, i.e., Algorithm~2, is labeled as solid red lines.  
The vertical bars represent the standard errors obtained from 100 replications.
In Fig.~\ref{simu_result}, we first observed that the classical dacKRR estimator, as expected, does not perform well.
We then observed that the proposed estimators perform consistently better than the classical estimator.
Such an observation indicates that when the response is highly skewed, oversampling the observations respecting the minority values of the response helps improve the estimation accuracy.
Finally, we observed that a larger value of $\tau$ is associated with a better performance. 
In particular, as the number of copies increases, the proposed estimator tends to have faster convergence rates.
This observation supports Corollary~\ref{coro}, which states that a larger number of observations after the oversampling process is associated with a smaller estimation MSE.

Besides Scott's rule ($l=O(N^{1/3})$),
we also considered other rules, e.g., the Sturge's formula ($l = O(\log_2N)$) \citep{sturges1926choice} and the Freedman-Diaconis choice ($l = O(N^{1/2})$) \citep{freedman1981histogram}.
Figure~\ref{simu_breaks} compares the empirical performance of the proposed estimator w.r.t. different choices of $l$'s.
For the setting of unskewed response (the first row), all methods share similar performance.
For response-skewed settings (the second and the third row), all the three aforementioned rules yield better performance than fixed $l$. 
Among all, Scott's rule gives the best results.
The simulation results are consistent with Corollary 4.2, which indicates that the proposed estimator shows advantages over the classical one. 

Besides the impact of $l$, we also studied the impact of $k$.
In addition, we have also applied the proposed strategy to other nonparametric regression estimators, say smoothing splines, under the divide-and-conquer setting. 
Such simulation results are provided in the Supplementary Material.
These results indicated that the performance of the proposed method is robust the choice of $k$ and has the potential to improve the estimation of other nonparametric regression estimators under the divide-and-conquer setting.
It is also possible to consider unequally-spaced slicing methods in the proposed method; however, such extensions are beyond the scope of this paper.

\section{REAL DATA ANALYSIS}



We applied the proposed method to a real-world dataset called Melbourne-housing-market \footnote{Data source https://www.kaggle.com/anthonypino/ melbourne-housing-market.}.
The dataset includes housing clearance data in Melbourne from the year 2016 to 2017.
Each observation represents a record for a sold house, including the total price and several predictors.
Of interest is to predict the price-per-square-meter (range from 20 to 30000) for each sold house using the longitude and the latitude of the house, and the distance from the house to the central business district (CBD).
Such a goal can be achieved by using kernel ridge regression.



\begin{figure}[!ht]
    \begin{center}
        \begin{tabular}{c}
            \includegraphics[scale = .45]{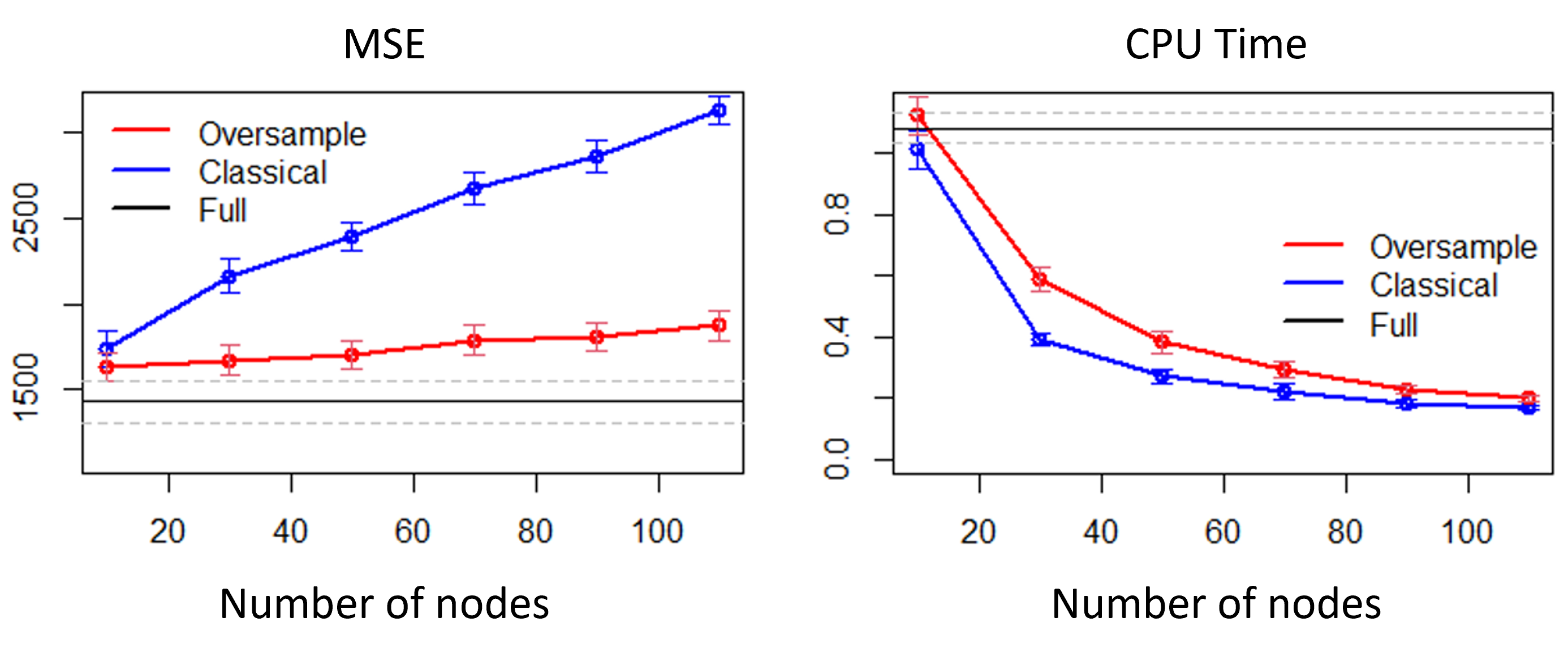}\\
        \end{tabular}
        \vspace{-5pt}
        \caption{Left panel: testing MSE versus different number of nodes. Right panel: CPU time (in seconds) versus different number of nodes. Vertical bars represent the standard errors.}\label{house_res}
    \end{center}
     \vspace{-15pt}
\end{figure}

We replicated the experiment one hundred times.
In each replicate, we used stratified sampling to randomly pick $10\%$ of the observations as the testing set and the remaining as the training set.
The Gaussian kernel function was used for the kernel ridge regression.
We set the number of nodes $k = \{10, 30, 50, 70, 90, 110\}$. 
We followed the procedure in \cite{zhang2015divide} to select the bandwidth for the kernel and the regularization parameters.
The performance of different estimators was evaluated by the MSE, respecting the prediction on the testing set.
The experiment was conducted in \texttt{R} using a Windows computer with 16GB of memory and a single-threaded 3.5Ghz CPU.

The left panel of Fig.~\ref{house_res} shows the MSE versus the different number of nodes $k$, and the right panel of Fig.~\ref{house_res} shows the CPU time (in seconds) versus different $k$.
To the best of our knowledge, our work is the first approach that addresses the response-skewed issue under the dacKRR setting, while other approaches only considered discrete-response settings or cannot be easily extended to the divide-and-conquer scenario. Thus we only compare our method with the classical dacKRR approach. In Fig.~\ref{house_res}, the blue lines and the red lines represent the classical dacKRR approach and the proposed approach, respectively.
Vertical bars represent the standard errors, which are obtained from one hundred replicates.
The black lines represent the results of the full sample KRR estimate, and the gray dashed lines represent its standard errors.
We can see that the proposed estimate consistently outperforms the classical dacKRR estimate. 
In particular, we observe the MSE of the proposed estimate is comparable with the MSE of the full sample KRR estimate.
The MSE for the classical dacKRR estimate, however, almost doubles the MSE for the full sample KRR estimate when the number of nodes $k$ is greater than one hundred.
The bottom panel of Fig.~\ref{house_res} shows the CPU time of the proposed approach is comparable with the CPU time of the classical divide-and-conquer approach.
All these observations are consistent with the findings in the previous section.
Such observations indicate the proposed approach outperforms the classical dacKRR approach for response-skewed data without requiring too much extra computing time.


\appendix
\section[A]{SIMULATION STUDIES}

\subsection{Impact of The Number of Nodes}

We considered the function as in the manuscript, i.e.,
\begin{equation*}
    g(\bm{x}, \bm{c}) = \frac{0.1}{\|\bm{x}-\bm{c}\| +0.05}\sin\left(\frac{0.01\pi}{\|\bm{x}-\bm{c}\|+0.05}\right),
\end{equation*}
where $\bm{x}\in [0,1]^d$, and $\|\cdot\|$ represents the $\mathbb{L}_2$ norm.
We then simulated the data from Model (1) with $N=10^4$, $d=1,2,4,6$, and two different regression function $\eta_0$'s,

{\textbf{Uni-peak}}: $\eta_0(\bm{x}) = g(\bm{x}, \bm{c})$, with $\bm{c} = 0.4\times\bm{1}^d$;

{\textbf{Double-peak}}: $\eta_0(\bm{x}) = g(\bm{x}, \bm{c}_1) + 0.4g(\bm{x}, \bm{c}_2)$, with $\bm{c}_1 = 0.4\times\bm{1}^d$ and $\bm{c}_2 = 0.7\times\bm{1}^d$.

\begin{figure}[!ht]
    \begin{center}
        \begin{tabular}{c}
            \includegraphics[scale = .5]{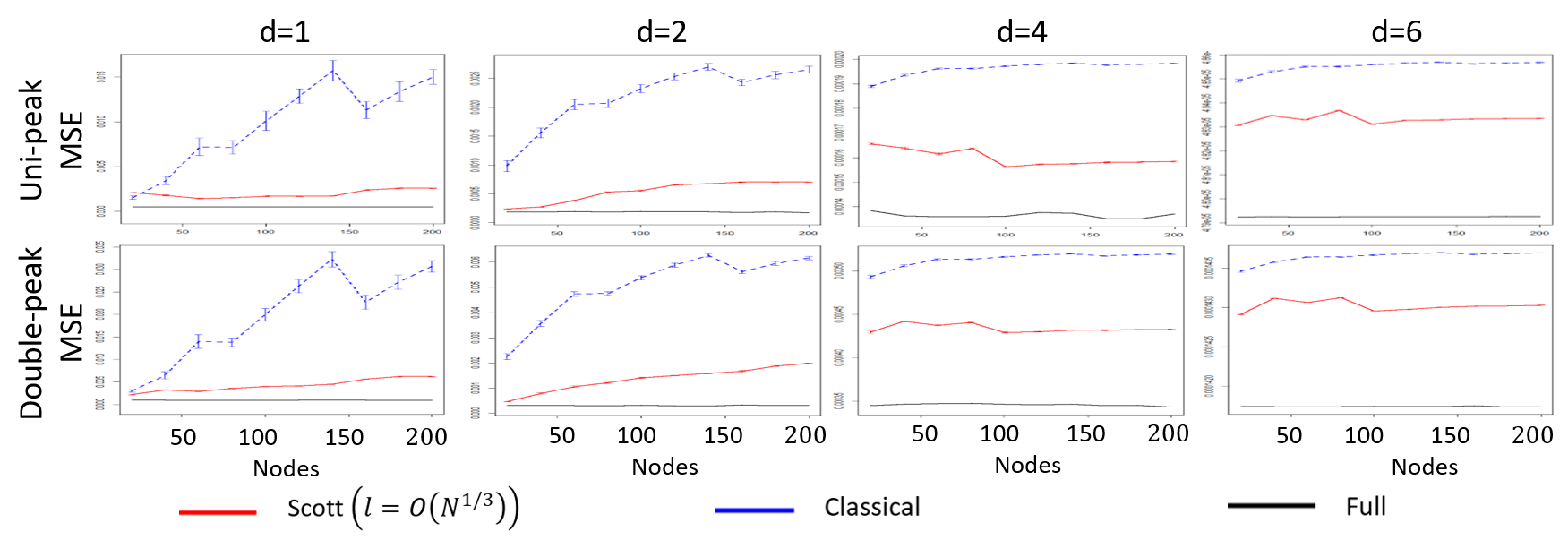}\\
        \end{tabular}
        \caption{Comparison for different choices of $k$'s.}\label{simu_nodes}
    \end{center}
\end{figure}

Figure~\ref{simu_nodes} study the impact of the number of nodes $k$.
In Fig.~\ref{simu_nodes}, we fix $N$ and let $k$ varies.
We observe that the proposed estimator consistently outperforms the classical one under all scenarios.

\subsection{Application in Other Non-parametric Regression Methods}

\begin{figure}[!ht]
    \begin{center}
        \begin{tabular}{c}
            \includegraphics[scale = .5]{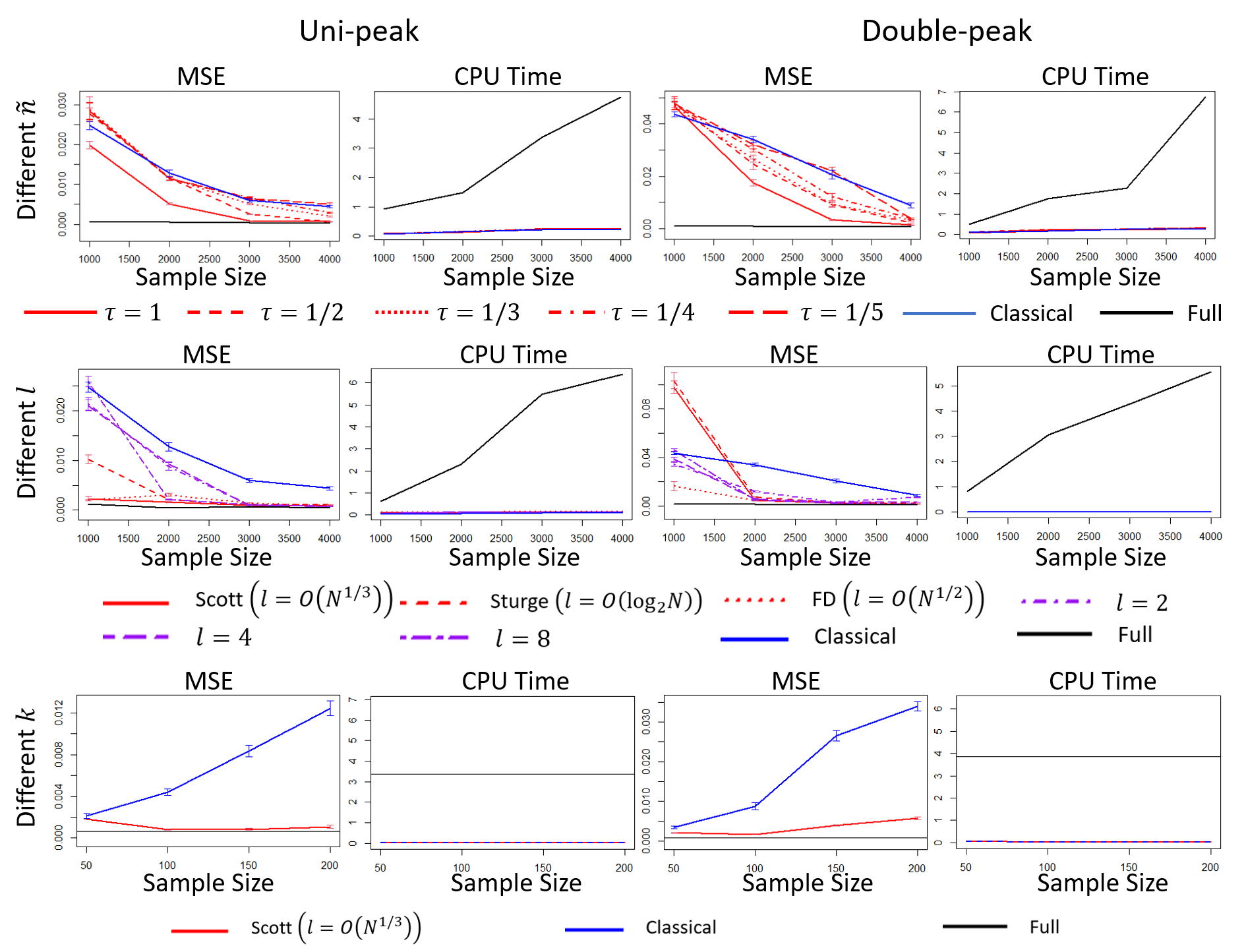}\\
        \end{tabular}
        \caption{Application of the proposed method in smoothing splines.}\label{ssanova}
    \end{center}
\end{figure}

Besides the KRR estimator, we also applied the proposed response-adaptive partition strategy to other non-parametric regression methods. One example is the smoothing splines under the divide-and-conquer setting. In the simulation, we again study the impact of the over-sample size $\tilde{n}$, the number of slices $l$, and the number of nodes $k$. We set the dimension $d=4$, other settings are similar to the cases for KRR. The results are shown in Fig.~\ref{ssanova}. Each row of Fig.~\ref{ssanova} shows the impact of $\tilde{n}, l$ and $k$, respectively. The odd columns represent the estimation MSE, and the even columns represent the CPU time in seconds. The first two columns represent the results of the uni-peak setting, and the last two columns represent the results of the double-peak setting. We can see that the results show the similar pattern as in the KRR cases. These results indicate that the performance of the proposed method is robust for different nonparametric regressions.

\section[B]{PROOFS}

\subsection{Proof of Theorem 4.1}
The proof of Theorem 4.1 relies on the following bias-variance decomposition
\begin{equation*} 
    \begin{split}
        \ex \{\norm{\bar{\eta}-\eta_0}^2_{\mathcal{P}_X}\} 
        & = \ex \{\norm{\ex(\bar{\eta})-\eta_0+\bar{\eta}-\ex(\bar{\eta})}^2_{\mathcal{P}_X}\}  \\
        & = \ex \{\norm{\ex(\bar{\eta})-\eta_0}^2_{\mathcal{P}_X}\} + \ex \Bigg\{\norm{\frac{1}{k}\sum_{j=1}^k(\hat{\eta}_j-\ex(\hat{\eta}_j))}^2_{\mathcal{P}_X}\Bigg\} \\
        & = \ex \{\norm{\ex(\hat{\eta}_1)-\eta_0}^2_{\mathcal{P}_X}\} + \frac{1}{k} \ex \{\norm{\hat{\eta}_1-\ex(\hat{\eta}_1)}^2_{\mathcal{P}_X}\},
    \end{split}
\end{equation*}
where $\ex\{\norm{\bar{\eta}-\eta_0}_{\mathcal{P}_X}^2\} = \ex\{\int_X (\bar{\eta}(\X_1)- \eta_0(\X_1))^2 d\mathcal{P}_X\}$. The claim of Theorem 4.1 can be proved by combining the inequalities in Lemma \ref{le:var} and \ref{le:bias}. The minimax bound for risk is
\begin{equation*}
  \begin{split}
    \ex \{\norm{\bar{\eta}-\eta_0}^2_{\mathcal{P}_X}\}  &\leq 8\lambda \norm{\eta_0}^2_{\mathcal{H}} + \frac{4d_{\lambda}\sigma^2}{\tilde{n}k}+ 4d_{\lambda}\\
    &\times(\mu_1^2 \norm{\eta_0}^2_{\mathcal{H}}+\frac{\omega^2\sigma^2}{\lambda \tilde{n}k})\exp\bigg(-\frac{3\lambda \tilde{n}}{28\omega^2}\bigg).
   \end{split}
\end{equation*}
With the properly chosen smoothing parameter, the first two terms in the upper bound are typically dominate for most of the commonly used kernels, e.g., the kernels for the Sobolev spaces, see  \cite{gu2013smoothing}. For any kernel with $\nu$-polynomial eigendecay, we obtain the optimal convergence rate for risk, i.e., $O((1/\tilde{N})^{2\nu/(2\nu+1)})$, with $\lambda = (1/\tilde{N})^{2\nu/(2\nu+1)}$ for the finite number of nodes. 
$\square$

\begin{lemma}
\label{le:var}
Under Assumption 1-3, we have 
\begin{equation*}
\ex \{\norm{\ex(\hat{\eta}_1)-\eta_0}^2_{\mathcal{P}_X}\} \leq 8\lambda \norm{\eta_0}^2_{\mathcal{H}} + 4d_{\lambda}\mu_1^2 \norm{\eta_0}^2_{\mathcal{H}}\exp\bigg(-\frac{3\lambda \tilde{n}}{28\omega^2}\bigg).
\end{equation*}
\end{lemma}
\noindent
\emph{Proof.} The estimate $\hat{\eta}_1$ minimizes the following empirical objective, 
\begin{equation}
    \mathcal{L}(\eta) = \frac{1}{2\tilde{n}}\sum_{i=1}^{\tilde{n}} (\tilde{y}_{1i}-\la K_{\X_{1i}}, \eta \ra_{\mathcal{H}})^2 + \frac{1}{2} \la J_{\lambda} \eta, \eta \ra_{\mathcal{H}},
\label{eq:frechet}
\end{equation}
where ($\tilde{y}_{1i}$, $\X_{1i}$), $i=1,\cdots,\tilde{n}$ are the data assigned to the first node, $J_{\lambda}: \mathcal{H} \rightarrow \mathcal{H}$ is a self-adjoint operator such that $\la J_{\lambda} \eta, \eta \ra_{\mathcal{H}} = \lambda \la \eta, \eta \ra_{\mathcal{H}}$ for $\eta \in \mathcal{H}$, and $\eta(\X_{1i})=\la K_{\X_{1i}}, \eta \ra_{\mathcal{H}}$. Note that the objective function \eqref{eq:frechet} is Fr\'echet differentiable \cite{luenberger1997optimization}. We can obtain 
\begin{equation*}
\begin{split}
    \frac{\partial \mathcal{L}(\eta)}{\partial \eta} & = -\frac{1}{\tilde{n}}\sum_{i=1}^{\tilde{n}} (\tilde{y}_{1i}-\la K_{\X_{1i}}, \eta\ra_{\mathcal{H}})K_{\X_{i1}} + \lambda \eta \\
    &= -\frac{1}{\tilde{n}}\sum_{i=1}^{\tilde{n}} \tilde{y}_{1i} K_{\X_{1i}} + \frac{1}{\tilde{n}}\sum_{i=1}^{\tilde{n}}\la K_{\X_{1i}}, \eta\ra_{\H} K_{\X_{1i}} + \lambda \eta.
\end{split}
\end{equation*}
For notation simplicity, we define 
\begin{equation*}
\begin{split}
    \Gamma_{\X_1} \eta = \frac{1}{\tilde{n}}\sum_{i=1}^{\tilde{n}} \la K_{\X_{1i}}, \eta \ra_{\H} K_{\X_{1i}} = \frac{1}{\tilde{n}}\sum_{i=1}^{\tilde{n}}\eta(\X_{1i})K_{\X_{1i}}, \\ 
    F_{\X_{1}}^* \bm{\tilde{y}}_1 = \frac{1}{\tilde{n}}\sum_{i=1}^{\tilde{n}} \tilde{y}_{1i} K_{\X_{1i}}, 
\end{split}
\end{equation*}
where $F_{\X_{1}}^*$ is the adjoint of $F_{\X_1} = (\la \eta, K_{\X_{11}}\ra_{\H}, \cdots, \la \eta, K_{\X_{1\tilde{n}}}\ra_{\H})^{\T} = (\eta(\X_{11}), \cdots, \eta(\X_{1\tilde{n}}))^{\T}$. 
We then have that 
\begin{equation*}
 \frac{\partial \mathcal{L}(\eta)}{\partial \eta} = - F_{\X_{1}}^* \bm{\tilde{y}}_1 + \Gamma_{\X_1} \eta + \lambda \eta,
\end{equation*}
which yields the estimate 
\begin{equation*}
\hat{\eta}_1 = (\Gamma_{\X_1}+\lambda I)^{-1} F_{\X_1}^* \bm{\tilde{y}}_1,
\end{equation*}
where $\bm{\tilde{y}}_1=(y_{11},\cdots,y_{1\tilde{n}})^{\T}$. Using the fact that $\bm{\tilde{y}}_{1}=F_{\X_{1}}\eta_0 + \bm{\epsilon}_{1}$, and plugging it into the equation above yields
\begin{equation}
\label{eq:b1}
    \hat{\eta}_1 = (\Gamma_{\X_1}+\lambda I)^{-1} F_{\X_1}^* F_{\X_1} \eta_0 + (\Gamma_{\X_1}+\lambda I)^{-1} F_{\X_1}^* \bm{\epsilon}_1.
\end{equation}
Taking expectation on both sides of \eqref{eq:b1}, we obtain 
\begin{equation*}
    \ex(\hat{\eta}_1) = (\Gamma_{\X_1}+\lambda I)^{-1} F_{\X_1}^* F_{\X_1} \eta_0.
\end{equation*}
The difference between $\ex(\hat{\eta}_1)$ and $\eta_0$ can be written as 
\begin{equation*}
    \eta_0 - \ex(\hat{\eta}_1) =  \{I - (\Gamma_{\X_1}+\lambda I)^{-1}\Gamma_{\X_1}\} \eta_0,
\end{equation*}
where $\Gamma_{\X_1} \eta_0 = \frac{1}{\tilde{n}}\sum_{i=1}^{\tilde{n}} \la K_{\X_{1i}}, \eta_0 \ra_{\H} K_{\X_{1i}} = \frac{1}{\tilde{n}}\sum_{i=1}^{\tilde{n}}\eta_0(\X_{1i})K_{\X_{1i}} = F_{\X_1}^* F_{\X_1}\eta_0$. Thus, we have
\begin{equation}
\label{eq:b2}
    \ex \{\norm{\ex(\hat{\eta}_1)-\eta_0}^2_{\mathcal{P}_X}\} = \ex\norm{\Gamma^{1/2}\{I - (\Gamma_{\X_1}+\lambda I)^{-1}\Gamma_{\X_1}\} \eta_0}_{\H}^2,
\end{equation}
where $\Gamma \eta = \int_X \la\eta, K_{\bm{x}} \ra_{\H}K_{\bm{x}}d\mathcal{P}_X(\bm{x})$ for $\eta \in \H$. The equation above uses the fact that one can move between $\H$ and $L^2(\mathcal{P}_X)$ inner products as follows
\begin{equation*}
    \la f, g \ra_{\mathcal{P}_X} = \la f, \Gamma g \ra_{\H}.
\end{equation*}
Let $\phi_k$, $k=1,2,\cdots$ be the orthonormal basis for $\H$ and $\bm{\Phi}=(\phi_k(\X_{1i}))_{1\le i \le \tilde{n}; 1\le k < \infty}$ be the $n\time \infty$ matrix based on the orthonormal basis. We then have $\eta_0=\sum_{k=1}^{\infty}a_k\phi_k$, and the norm in \eqref{eq:b2} can be written as 
\begin{equation*}
     \ex\norm{\bm{\Gamma}^{1/2}\{\bm{I} - (\bm{\Gamma}_{\tilde{n}}+\lambda \bm{I})^{-1}\bm{\Gamma}_{\tilde{n}}\} \bm{a}}_{\ell^2(\mathbb{N})}^2,
\end{equation*}
where $\bm{\Gamma}=\text{diag}(\mu_1^2,\mu_2^2,\cdots)$, $\bm{I}=\text{diag}(1,1,\cdots)$, $\bm{\Gamma}_{\tilde{n}}=\frac{1}{n}\bm{\Phi}^{\T}\bm{\Phi}$, and $\bm{a}=(a_1,a_2,\cdots)^{\T}$.
\begin{equation}
\label{eq:b3}
\begin{split}
    \ex\norm{\bm{\Gamma}^{1/2}\{\bm{I} - (\bm{\Gamma}_{\tilde{n}}+\lambda \bm{I})^{-1}\bm{\Gamma}_{\tilde{n}}\} \bm{a}}_{\ell^2(\mathbb{N})}^2
    &\leq \ex \bigg\{\norm{\bm{\Gamma}^{1/2}\{\bm{I} - (\bm{\Gamma}_{\tilde{n}}+\lambda \bm{I})^{-1}\bm{\Gamma}_{\tilde{n}}\}}^2_{\ell^2(\mathbb{N})} \mathbbm{1}(\mathcal{C}_z^c)\bigg\} \norm{\bm{a}}^2_{\ell^2(\mathbb{N})} \\ 
    &+ \ex \bigg\{\norm{\bm{\Gamma}^{1/2}\{\bm{I} - (\bm{\Gamma}_{\tilde{n}}+\lambda \bm{I})^{-1}\bm{\Gamma}_{\tilde{n}}\}}^2_{\ell^2(\mathbb{N})} \mathbbm{1}(\mathcal{C}_z) \bigg\} \norm{\bm{a}}^2_{\ell^2(\mathbb{N})},
\end{split}
\end{equation}
where the event for $0<z<1$,
\begin{equation*}
    \mathcal{C}_z = \Bigg\{\norm{(\bm{\Gamma}+\lambda \bm{I})^{-1/2}(\bm{\Gamma}_{\tilde{n}}-\bm{\Gamma})(\bm{\Gamma}+\lambda \bm{I})^{-1/2})}_{\ell^2(\mathbb{N})} \ge z\Bigg\},
\end{equation*}
and $\mathcal{C}_z^c$ is the complement of $\mathcal{C}_z$. We use the fact that $\mathbbm{1}(\mathcal{C}_z^c)*\mathbbm{1}(\mathcal{C}_z)=0$ to obtain the inequality in \eqref{eq:b3}. 
We now bound the first term on the right-hand side of \eqref{eq:b3}. 
\begin{equation*}
\begin{split}
   \norm{\bm{\Gamma}^{1/2}\{\bm{I} - (\bm{\Gamma}_{\tilde{n}}+\lambda \bm{I})^{-1}\bm{\Gamma}_{\tilde{n}}\}}^2_{\ell^2(\mathbb{N})} &= \norm{\bm{\Gamma}^{1/2}(\bm{\Gamma}_{\tilde{n}}+\lambda \bm{I})^{-1} (\bm{\Gamma}_{\tilde{n}}+\lambda \bm{I}) (\bm{I} - (\bm{\Gamma}_{\tilde{n}}+\lambda \bm{I})^{-1}\bm{\Gamma}_{\tilde{n}})}^2_{\ell^2(\mathbb{N})} \\
   & \leq \norm{(\bm{\Gamma}_{\tilde{n}}+\lambda \bm{I})^{-1/2}}^2_{\ell^2(\mathbb{N})} \norm{(\bm{\Gamma}_{\tilde{n}} + \lambda \bm{I}) (\bm{I} - (\bm{\Gamma}_{\tilde{n}}+\lambda \bm{I})^{-1}\bm{\Gamma}_{\tilde{n}})}^2_{\ell^2(\mathbb{N})} \norm{\bm{\Gamma}^{1/2}(\bm{\Gamma}_{\tilde{n}}+\lambda \bm{I})^{-1/2}}^2_{\ell^2(\mathbb{N})} \\
   &\leq 4\lambda^2 \norm{\bm{\Gamma}(\bm{\Gamma}_{\tilde{n}}+\lambda \bm{I})^{-1}}_{\ell^2(\mathbb{N})} \norm{(\bm{\Gamma}_{\tilde{n}}+\lambda \bm{I})^{-1}}_{\ell^2(\mathbb{N})}. 
\end{split}
\end{equation*}
The last inequality follows based on Definition 1 in \cite{BAUER200752}. 
For $0\leq \gamma \leq 1$, we have 
\begin{equation*}
    \begin{split}
      \norm{\bm{\Gamma}^{\gamma/2}(\bm{\Gamma}_{\tilde{n}}+\lambda \bm{I})^{-1}\bm{\Gamma}^{\gamma/2}}_{\ell^2(\mathbb{N})} &= \norm{\bm{\Gamma}^{\gamma/2}\{(\bm{\Gamma}_{\tilde{n}}-\bm{\Gamma})+(\bm{\Gamma}+\lambda \bm{I})\}^{-1}\bm{\Gamma}^{\gamma/2}}_{\ell^2(\mathbb{N})} \\
      & \leq \norm{\bm{\Gamma}^{\gamma}(\bm{\Gamma}+\lambda \bm{I})^{-1}}_{\ell^2(\mathbb{N})}\norm{(\bm{I}-(\bm{\Gamma}+\lambda \bm{I})^{-1/2} (\bm{\Gamma}-\bm{\Gamma}_{\tilde{n}})(\bm{\Gamma}+\lambda \bm{I})^{-1/2})^{-1}}_{\ell^2(\mathbb{N})} \\
      & \leq \lambda^{\gamma-1} \norm{(\bm{I}-(\bm{\Gamma}+\lambda \bm{I})^{-1/2} (\bm{\Gamma}-\bm{\Gamma}_{\tilde{n}})(\bm{\Gamma}+\lambda \bm{I})^{-1/2})^{-1}}_{\ell^2(\mathbb{N})},
    \end{split}
\end{equation*}
where the last inequality follows based on the spectral theorem. Detailed discussions can be found in \cite{caponnetto2007optimal} and \cite[Chapter~9]{gu2013smoothing}. Applying the above inequality on the event $\mathcal{C}_z^{c}$ when $\gamma=1$ and $\gamma=0$, we obtain
\begin{equation*}
    \norm{\bm{\Gamma}^{1/2}\{\bm{I} - (\bm{\Gamma}_{\tilde{n}}+\lambda \bm{I})^{-1}\bm{\Gamma}_{\tilde{n}}\}}^2_{\ell^2(\mathbb{N})} \mathbbm{1}(\mathcal{C}_z^c) \leq \frac{4\lambda}{(1-z)^2},
\end{equation*}
for any $0<z \leq 1/2$. 

For the second term in \eqref{eq:b3}, we have that 
\begin{equation*}
\begin{split}
    \ex \bigg\{\norm{\bm{\Gamma}^{1/2}\{\bm{I} - (\bm{\Gamma}_{\tilde{n}}+\lambda \bm{I})^{-1}\bm{\Gamma}_{\tilde{n}}\}}^2_{\ell^2(\mathbb{N})} \mathbbm{1}(\mathcal{C}_z) \bigg\} \norm{\bm{a}}^2_{\ell^2(\mathbb{N})} &\leq \norm{\bm{\Gamma}^{1/2}}_{\ell^2(\mathbb{N})}^2\norm{\bm{I} - (\bm{\Gamma}_{\tilde{n}}+\lambda \bm{I})^{-1}\bm{\Gamma}_{\tilde{n}}}_{\ell^2(\mathbb{N})}^2P(\mathcal{C}_z) \norm{\bm{a}}^2_{\ell^2(\mathbb{N})} \\
    & \leq \mu_1^2 \norm{\bm{a}}^2_{\ell^2(\mathbb{N})} P(\mathcal{C}_z),
\end{split}
\end{equation*}
where the last inequality follows based on Definition 1 in \cite{BAUER200752}.  
We complete the proof by applying Lemma 1 adapted from \cite{dicker2017kernel}, which states that
\begin{equation*}
    P(\mathcal{C}_z) \leq 4d_{\lambda} \exp\bigg(-\frac{3\lambda \tilde{n}}{28\omega^2}\bigg).
\end{equation*}
$\square$

\begin{lemma}
\label{le:bias}
Under Assumption 1-3, we have  
\begin{equation*}
    \ex\{\norm{\hat{\eta}_1-\ex(\hat{\eta}_1)}^2_{\mathcal{P}_X}\} \leq \frac{4d_{\lambda}\sigma^2}{\tilde{n}} + 4d_{\lambda}\frac{\omega^2\sigma^2}{\lambda \tilde{n}} \exp(-\frac{3\lambda\tilde{n}}{28\omega^2}).
\end{equation*}
\end{lemma}
\noindent
\emph{Proof.}  Note that 
\begin{equation*}
    \ex(\hat{\eta}_1) = (\Gamma_{\X_1}+\lambda I)^{-1} F_{\X_1}^* F_{\X_1} \eta_0.
\end{equation*}
From \eqref{eq:b1}, we thus have 
\begin{equation}
\label{eq:var}
    \ex\{\norm{\hat{\eta}_1-\ex(\hat{\eta}_1)}^2_{\mathcal{P}_X}\} = \ex\norm{\Gamma^{1/2}(\Gamma_{\X_1}+\lambda I)^{-1} F_{\X_1}^* \bm{\epsilon}_1}^2_{\mathcal{H}}.
\end{equation}
Following similar techniques in Lemma 2.1, the norm of $\H$ in \eqref{eq:var} can be written as 
\begin{equation*}
    \ex\norm{\Gamma^{1/2}(\Gamma_{\X_1}+\lambda I)^{-1} F_{\X_1}^* \bm{\epsilon}_1}^2_{\mathcal{H}} = \frac{1}{\tilde{n}^2}\ex\norm{\bm{\Gamma}^{1/2}(\bm{\Gamma}_{\tilde{n}}+\lambda \bm{I})^{-1}\bm{\Phi}^{\T}\bm{\epsilon}_1}^2_{\ell^2(\mathbb{N})}.
\end{equation*}
By Von Neumann's inequality and the assumption that $\ex[\epsilon_{1i}^2] = \sigma^2$ for $i=1,\cdots,\tilde{n}$, we obtain 
\begin{equation*}
    \frac{1}{\tilde{n}^2}\ex\norm{\bm{\Gamma}^{1/2}(\bm{\Gamma}_{\tilde{n}}+\lambda \bm{I})^{-1}\bm{\Phi}^{\T}\bm{\epsilon}_1}^2_{\ell^2(\mathbb{N})} \leq \frac{\sigma^2}{\tilde{n}} \ex \{\Tr (\bm{\Gamma}(\bm{\Gamma}_{\tilde{n}}+\lambda \bm{I})^{-2}\bm{\Gamma}_{\tilde{n}}) \}.
\end{equation*}
The rest of proof is similar to that in Lemma 2.2. Following the same technique in \eqref{eq:b3}, we have 
\begin{equation}
\label{eq:main2}
\begin{split}
    \frac{1}{\tilde{n}^2}\ex\norm{\bm{\Gamma}^{1/2}(\bm{\Gamma}_{\tilde{n}}+\lambda \bm{I})^{-1}\bm{\Phi}^{\T}\bm{\epsilon}_1}^2_{\ell^2(\mathbb{N})} &\leq  \frac{\sigma^2}{\tilde{n}}\Tr(\bm{\Gamma}(\bm{\Gamma}_{\tilde{n}}+\lambda \bm{I})^{-2}\bm{\Gamma}_{\tilde{n}}) P(\mathcal{C}_z)\\
    & + \frac{\sigma^2}{\tilde{n}}\ex \{\Tr (\bm{\Gamma}(\bm{\Gamma}_{\tilde{n}}+\lambda \bm{I})^{-2}\bm{\Gamma}_{\tilde{n}}) \mathbbm{1}(\mathcal{C}_z^c)\}.
\end{split}
\end{equation}
We now bound the first term on the right-hand side of \eqref{eq:main2}. By Von Neumann inequality, we have 
\begin{equation}
    \label{eq:b4}
    \Tr(\bm{\Gamma}(\bm{\Gamma}_{\tilde{n}}+\lambda \bm{I})^{-2}\bm{\Gamma}_{\tilde{n}}) \leq \Tr(\bm{\Gamma}) \norm{\bm{\Gamma}_{\tilde{n}}+\lambda \bm{I})^{-2}\bm{\Gamma}_{\tilde{n}})}_{\ell^2(\mathbb{N})}.
\end{equation}
From Cauchy-Schwartz inequality, we have the following inequality for any $\eta \in \H$
\begin{equation*}
    \la \Gamma_{\X_1} (\Gamma_{\X_1}+\lambda I)^{-1}\eta,  (\Gamma_{\X_1}+\lambda I)^{-1} \eta \ra_{\H} \leq \norm{\Gamma_{\X_1} (\Gamma_{\X_1}+\lambda I)^{-1}\eta}_{\H} \norm{(\Gamma_{\X_1}+\lambda I)^{-1} \eta}_{\H}. 
\end{equation*}
From Definition 1 in \cite{BAUER200752}, we also have 
\begin{equation*}
    \norm{\Gamma_{\X_1} (\Gamma_{\X_1}+\lambda I)^{-1}\eta}_{\H} \leq \norm{\eta}_{\H}, 
\end{equation*}
and 
\begin{equation*}
    \norm{(\Gamma_{\X_1}+\lambda I)^{-1} \eta}_{\H} \leq \frac{1}{\lambda}\norm{\eta}_{\H}.
\end{equation*}
Thus, we have 
\begin{equation}
\label{eq:b5}
    \norm{\bm{\Gamma}_{\tilde{n}}+\lambda \bm{I})^{-2}\bm{\Gamma}_{\tilde{n}})}_{\ell^2(\mathbb{N})} \leq \frac{1}{\lambda}. 
\end{equation}
Plugging \eqref{eq:b5} into \eqref{eq:b4}, we have 
\begin{equation*}
    \Tr(\bm{\Gamma}(\bm{\Gamma}_{\tilde{n}}+\lambda \bm{I})^{-2}\bm{\Gamma}_{\tilde{n}}) \leq \frac{\omega^2}{\lambda}
\end{equation*}
For the second term on the right-hand side of \eqref{eq:main2}, we have 
\begin{equation*}
\begin{split}
    \Tr (\bm{\Gamma}(\bm{\Gamma}_{\tilde{n}}+\lambda \bm{I})^{-2}\bm{\Gamma}_{\tilde{n}}) &= \Tr (\bm{\Gamma}(\bm{\Gamma}_{\tilde{n}}+\lambda\bm{I})^{-1}(\bm{\Gamma}_{\tilde{n}}+\lambda\bm{I})(\bm{\Gamma}_{\tilde{n}}+\lambda\bm{I})^{-2}\bm{\Gamma}_{\tilde{n}}) \\
    & \leq \Tr (\bm{\Gamma}(\bm{\Gamma}_{\tilde{n}}+\lambda\bm{I})^{-1}\norm{(\bm{\Gamma}_{\tilde{n}}+\lambda\bm{I})(\bm{\Gamma}_{\tilde{n}}+\lambda\bm{I})^{-2}\bm{\Gamma}_{\tilde{n}}}_{\ell^2_{\mathbb{N}}}, 
\end{split}
\end{equation*}
where the last inequality follows by Von Neumann's inequality. From \eqref{eq:b5} and Definition 1 in \cite{BAUER200752}, it is easy to verify that 
\begin{equation*}
    \norm{(\bm{\Gamma}_{\tilde{n}}+\lambda\bm{I})(\bm{\Gamma}_{\tilde{n}}+\lambda\bm{I})^{-2}\bm{\Gamma}_{\tilde{n}}}_{\ell^2_{\mathbb{N}}} \leq 2.
\end{equation*}
Applying Von Neumann's inequality again, we have 
\begin{equation*}
\Tr (\bm{\Gamma}(\bm{\Gamma}_{\tilde{n}}+\lambda \bm{I})^{-2}\bm{\Gamma}_{\tilde{n}}) \leq 2 d_{\lambda} \norm{(\bm{I}-(\bm{\Gamma}+\lambda \bm{I})^{-1/2} (\bm{\Gamma}-\bm{\Gamma}_{\tilde{n}})(\bm{\Gamma}+\lambda \bm{I})^{-1/2})^{-1}}_{\ell^2(\mathbb{N})},
\end{equation*}
which is bounded by $2d_{\lambda}/(1-z)$ on the event $\mathcal{C}_z^c$. We complete the proof by setting $z=1/2$ and using the fact that 
\begin{equation*}
    P(\mathcal{C}_z) \leq 4d_{\lambda} \exp\bigg(-\frac{3\lambda \tilde{n}}{28\omega^2}\bigg).
\end{equation*}
$\square$

\newpage
\bibliographystyle{abbrv}
\bibliography{ref}

\end{document}